\documentclass[conference]{IEEEtran}
\IEEEoverridecommandlockouts
\usepackage{amsmath,amssymb,amsfonts}
\usepackage{algorithmic}
\usepackage{graphicx}
\usepackage{textcomp}
\usepackage{xcolor}
\def\BibTeX{{\rm B\kern-.05em{\sc i\kern-.025em b}\kern-.08em
    T\kern-.1667em\lower.7ex\hbox{E}\kern-.125emX}}

\usepackage{csquotes}
\usepackage{multirow}
\usepackage{booktabs}
\newcommand{\ra}[1]{\renewcommand{\arraystretch}{#1}}
\usepackage{caption} 
\usepackage[normalem]{ulem}
\usepackage[
backend=biber,
style=numeric,
style=ieee,
sorting=ynt
]{biblatex}
\def\etal{\emph{et al.}}
\def\eg{\emph{e.g.}}

\begin{document}

\title{Learning structure-aware semantic segmentation with image-level supervision\\
}

\author{\IEEEauthorblockN{1\textsuperscript{st} Jiawei Liu, 2\textsuperscript{nd} Jing Zhang, 3\textsuperscript{rd} Yicong Hong, 4\textsuperscript{th} Nick Barnes}
\IEEEauthorblockA{\textit{College of Engineering and Computer Science} \\
\textit{Australian National University}\\
Canberra, Australia \\
\{jiawei.liu3, jing.zhang, yicong.hong, nick.barnes\}@anu.edu.au}
}

\maketitle

\begin{abstract}
Compared with expensive pixel-wise annotations, image-level labels make it possible to learn semantic segmentation in a weakly-supervised manner. Within this pipeline, the class activation map (CAM) is obtained and further processed to serve as a pseudo label to train the semantic segmentation model in a fully-supervised manner. In this paper, we argue that the lost structure information in CAM limits its application in downstream semantic segmentation, leading to deteriorated predictions. Furthermore, the inconsistent class activation scores
inside the same object contradicts the common sense that each region of the same object should belong to the same semantic category. To produce sharp prediction with structure information, we introduce an auxiliary semantic boundary detection module, which penalizes the deteriorated predictions. Furthermore, we adopt smoothness loss to encourage prediction inside the object to be consistent. Experimental results on the PASCAL-VOC dataset illustrate the effectiveness of the proposed solution.
\end{abstract}


\section{Introduction}
As a fundamental computer vision task, semantic segmentation aims to produce pixel-level classification of the given image. The advent of Convolutional Neural Networks (CNNs) and its subsequent advancements have significantly improved the performance of semantic segmentation models \cite{long2015fully, chen2017deeplab, zhou2016learning, chen2017rethinking}. However, the success of these models highly relies on a large training dataset with pixel-level annotations \cite{everingham2010pascal, cordts2016cityscapes}, which is expensive and time-consuming to obtain. To relieve the burden of pixel-wise labeling, weakly supervised semantic segmentation (WSSS) models have been widely studied, which are usually built upon weak annotations, including image-level labels \cite{wang2020self, fan2020learning, wei2018revisiting, ahn2018learning, huang2018weakly, araslanov2020single}, point supervisions \cite{bearman2016s}, scribble annotations \cite{vernaza2017learning, lin2016scribblesup} and bounding boxes \cite{song2019box, dai2015boxsup}. Among these weak annotations, image-level labels provide the lowest level of supervision, thus presenting the most challenging task.

\begin{figure}[htb]
    \centering
    \includegraphics[width=\columnwidth]{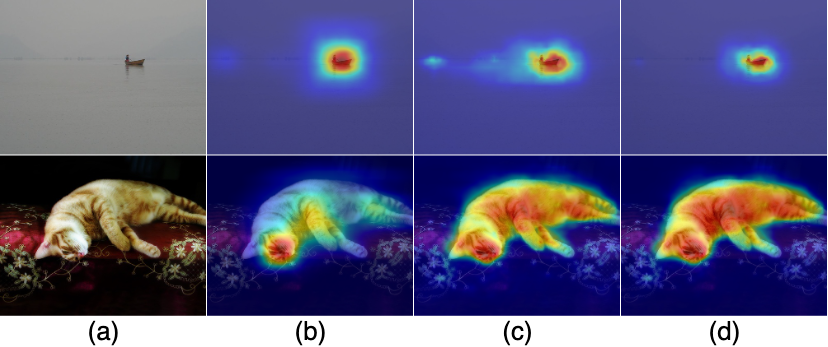}
    \vspace*{-0.5cm}
    \caption{\small 
    Comparison of class activation maps generated by (b) the basic CAM method \cite{zhou2016learning}, (c) SEAM \cite{wang2020self}, and (d) our proposed model.
    Our structure-aware class activation map leads to more consistently high activation within the object (row 2), and less spurious activation outside the object (row 1).
    }
    \label{fig: figure1}
\end{figure}

Most of the image-level label based WSSS models rely on a Class Activation Map (CAM) \cite{zhou2016learning} to provide initial localisation of the object. However, CAMs are incomplete, 
focusing on the most discriminative area and 
often revealing only a small part of the object, as shown in Fig.~\ref{fig: figure1} (b), which makes it insufficient to serve as 
a pseudo label for the downstream semantic segmentation task.

Recent studies \cite{ahn2018learning, wang2020self, chang2020weakly} have resorted to a three-stage training scheme to further boost the performance of the image-level supervision based WSSS models. Firstly, a better attention map is obtained to cover a larger part of the object, \eg, \cite{wang2020self, chang2020weakly} achieve this by incorporating additional constraints in the learning of a classification network. Secondly, the attention map is expanded through pixel correlations to increase the coverage rate. This is mainly achieved by learning an affinity matrix through 
AffinityNet \cite{ahn2018learning} to 
a
perform a random walk on the obtained class activation maps. Lastly, a semantic segmentation model, \eg \space DeepLab \cite{chen2017deeplab}, is trained using the expanded attention maps as pseudo labels in a fully supervised manner.

The performance of the three-stage training scheme highly depends on the quality of the initial class activation map. We argue that although the discriminative area is beneficial for object classification, the lost structure information in the class activation map hinders its performance in subsequent processing, \eg \space semantic segmentation.
Further, due to different degrees of discrimination, regions inside the same object have different class activation scores. In this paper, we propose a smoothing branch that leverages semantic boundary information to impose structural constraints on the training of WSSS models. It forces the resulting class activation maps to pay attention to object structures.  Furthermore, we introduce a semantic boundary-guided smoothness loss that achieves a more consistent attention map within the same object area while ignoring the pixels belonging to other classes as shown in Fig.~\ref{fig: figure1}. With the proposed framework, we can achieve better CAMs with structure information, leading to relatively sharp semantic predictions as shown in Fig. \ref{fig: figure1}.

Our main contributions can be summarised as:
\begin{enumerate}
    \item We propose a smoothing branch that leverages semantic boundaries.
    for weakly supervised semantic segmentation to obtain structure preserving attention maps.
    \item We present a semantic boundary-guided smoothness loss function that enforces more consistent semantic prediction
    within the same object area.
    \item Our proposed method achieves the state-of-the-art results on the PASCAL VOC 2012 dataset.
\end{enumerate}

\section{Related Work}
\subsection{Weakly supervised semantic segmentation}
Weakly supervised semantic segmentation models aim to perform the semantic segmentation task with only coarse annotations. Successful examples have explored the potential of bounding box \cite{song2019box, dai2015boxsup}, scribble \cite{vernaza2017learning, lin2016scribblesup}, point \cite{bearman2016s} and image-level \cite{wang2020self, fan2020learning, wei2018revisiting, ahn2018learning, huang2018weakly, araslanov2020single} annotations. Among these, 
image-level labels require the minimum-level of manual annotation
Further,
they present the most challenging task, providing the least supervision for training. Our work employs image-level labels as weak supervision for semantic segmentation task.

\subsection{Class Activation Maps}
Recent weakly supervised semantic segmentation works have been based on using CAMs \cite{zhou2016learning} to provide the initial localisation cues for different classes. It requires modification to the structure of CNNs trained for a classification task with image-level labels. The penultimate global average pooling (GAP) layer is removed, and the classification layer is directly applied to the high dimensional features to produce the pixel-level score maps for each class. Grad-CAM \cite{selvaraju2017grad} aims to mitigate the limitation of CAM that it is only applicable to fully convolutional neural networks and its performance is slightly reduced due to the change of network structure. It associates the importance of feature maps at the penultimate layer with the class gradients and computes the score maps with a weighted sum of those feature maps. However, these class activation maps reveal only the most discriminative parts of the objects as they are trained primarily for classification. The output pseudo segmentation maps are incomplete and require further refinement.

Some research approaches enforce the network to learn a more complete representation by erasing the most discriminative parts of the object. Wei \etal \cite{wei2017object} proposes to \enquote{erase} the obtained high-confidence area and retrain the network for the classification task with partially erased inputs in order to enforce
the network to identify more activation areas associated with each class. However, as a result of the gradual removal of the high confidence area, the model eventually includes incorrect pixels surrounding the target object for classification. To mitigate this problem, SeeNet \cite{hou2018self} employs two additional decoder branches to confine the erasure and attention within the object. DCSP \cite{chaudhry2017discovering} applies the erasure iteratively and leverages an off-the-shelf saliency detector to confine the process, so that the attention within the object can be accumulated.

Recent works have been focusing on improving the quality of class activation maps directly. FickleNet \cite{lee2019ficklenet} forces the model to learn a more complete CAM by randomly dropping connections of units in the convolution kernel. However, it is computationally expensive, involving generating 200 localisation maps for each input image and accumulating them to produce the revised result. OAA \cite{jiang2019integral} discovers a shift and transformation of CAMs in the training of the classification task. It accumulates CAMs in different stages of training to produce a final activation map that covers a larger area. Both SSENet \cite{wang2019self} and SEAM \cite{wang2020self} adopt scale equivariant attention to achieve self supervision. \cite{chang2020weakly} proposes to perform fine-grained classification within each class, and accumulate the sub-class activation maps with the class activation maps to enlarge the activation area. \cite{fan2020learning} learns a directional vector within each class to differentiate foreground and background pixels. However, these methods focus only on increasing the overlap ratio between the recovered and ground truth segmentation maps, while ignoring the structures of the objects, resulting in incorrect segmentation results around the object boundaries.

\begin{figure*}[htb]
    \centering
    \includegraphics[width=1.90\columnwidth]{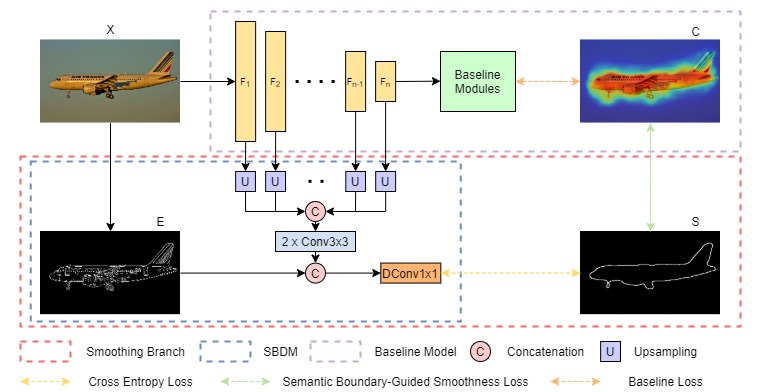}
    \vspace*{0cm}
    \caption{Our network consists of a baseline model \cite{wang2020self} and a smoothing branch. The Semantic Boundary Detection Module utilises feature maps and edge maps (\enquote{E}) obtained by applying 
    the Canny Edge detector on the input image (\enquote{X}) to make semantic boundary prediction. It encodes 
    boundary information into the output class activation map (\enquote{C}) by making feature maps structure-aware. The proposed semantic boundary guided smoothness loss further regulates the output CAM by enforcing consistently high activation scores within the object while creating a sharp score contrast along the semantic boundary (\enquote{S}).}
    \label{fig: Network}
\end{figure*}

\subsection{Structure-aware semantic segmentation}
Mismatch between the boundaries of CAMs and class objects is a key factor that hinders the recovery of the full segmentation mask of weakly supervised semantic segmentation. The idea of expanding the localisation cues, such as CAMs, to recover more complete representations of holistic objects was first proposed in \cite{kolesnikov2016seed}. The expansion can benefit from more accurate initial localisation \cite{li2018tell}. \cite{huang2018weakly} incorporates the seeded region growing (SDRG) algorithm into the network to expand the initial CAMs. The expansion is confined by the boundary that separates the foreground and background pixels. AffinityNet \cite{ahn2018learning} learns an affinity matrix on the initial CAMs and perform a random walk to diffuse confident labels to similar pixels. SSDD \cite{shimoda2019self} optimises the segmentation results of semantic segmentation models by learning from the dense Conditional Random Fields (dCRF) and Random Walk (RW) revised results. These works exploit pixel correlations to alleviate the mismatch between the segmentation map boundary and the class boundary, while some other works \cite{lee2019ficklenet, huang2018weakly, hou2018self, chaudhry2017discovering} directly adopt off-the-shelf saliency detection networks to localise the boundary between foreground and background pixels, which implicitly uses extra ground truth. Our work derives the semantic boundary from the semantic segmentation results of existing weakly supervised semantic segmentation model so that it does not implicitly or explicitly utilise any additional supervision. It utilises the semantic boundary to incorporate the structure information into the feature maps and make the activation score within the object more consistent, leading to improved segmentation results.

\section{Our Method}
In this section, we introduce the proposed structure-aware semantic segmentation model with image-level supervision as shown in Fig. \ref{fig: Network}.

\subsection{Smoothing Branch (SB)}
We adopt an existing WSSS model \cite{wang2020self} as our baseline model. Our proposed smoothing branch is implemented upon the baseline model by connecting to its encoder and CAM output layer without modification to its network structure. The convolutional layers producing feature maps of the same resolutions are grouped in different stages of network, and we denote the feature map produced by the last convolutional layer of each stage as $F_{m}(m = 1, \dots, n)$, where $n$ depends on the backbone used by the base model. The input image and image-level labels are denoted as $X$ and $Y$. The output CAMs from the base model are denoted as $C$. The number of classes in the dataset is defined as $k$.\\

\noindent\textbf{Semantic Boundary Detection Module (SBDM)}: The Semantic Boundary Detection Module encourages the feature maps of the baseline model to encode richer structural information. We use feature maps from different levels of the backbone network to produce semantic boundary predictions $B$ of size $(k + 1) \times W \times H$ corresponding to 20 classes in PASCAL VOC 2012 \cite{everingham2010pascal} plus the background, where $W \times H$ is the spatial size of the feature map. We define the upsampling operation for the $F_i$ as: 
$F_m^{U} = U (F_m)$, 
where $U$ applies a $1 \times 1$ convolution followed by 
a $3 \times 3$ convolution to channel size $M=32$, followed by an bilinear interpolation operation to achieve a feature map of the image size $W \times H$. We concatenate features of $F_{m}(m = 1, \dots, n)$, and feed it to another two $3 \times 3$ convolutional layers to obtain the initial semantic boundary of $k$ channels. 

To inform the model of the image structure we use handcrafted feature-based boundaries as per  \cite{takikawa2019gated,zhang2020weakly}. The computed Canny edges are expanded to $k$ channels before being interleaved between the initial semantic boundary through concatenation, Then they are fed to another $1 \times 1$ depthwise convolutional layer to obtain our final semantic boundary $B$. The whole process to obtain our final semantic boundary is shown in Eq. \ref{semantic_boundary_eq}:

\begin{equation}
\label{semantic_boundary_eq}
\begin{aligned}
B = DConv1(\amalg_C \{Conv3(Conv3(\amalg_C \{F_i^U\})), E\})
\end{aligned}
\end{equation}

\noindent
where $B$ is semantic boundary prediction, $\amalg_C{,}$ is concatenation operation, $Conv3$ is $3\times 3$ convolutional layer, $DConv1$ is $1 \times 1$ depthwise convolutional layer, and $E$ represents the output of applying the Canny edge detector on the input image. The SBDM is trained with a cross-entropy loss $\mathcal{L}_{B}$ with the preprocessed semantic boundary $S$ as supervision, which is derived from the final semantic segmentation results of our baseline \cite{wang2020self}.
\begin{equation}
    \mathcal{L}_{B} = \sum_{i=1}^{k} \sum_{u,v}^{W,H} (B_{u,v}^{i} \log (S_{u,v}^{i}) + (1 - B_{u,v}^{i}) \log (1 - S_{u,v}^{i})),
\end{equation}

\noindent\textbf{Semantic Boundary Guided Smoothness Loss:} We enforce the pixels inside the object to achieve consistently high activation scores and a sharp contrast to the pixels falling outside the boundary in terms of confidence scores. Inspired by \cite{wang2018occlusion, godard2017unsupervised}, we develop a semantic boundary guided smoothness loss to impose such a constraint. The original smoothness loss is designed to smooth the intensity while preserving structure information in the whole image. However, this is detrimental to the semantic segmentation task where structure information other than the class boundary needs to be suppressed. 
Further, 
this becomes even worse for WSSS with only image labels. The smoothness loss function interprets the edges as gradients of class activation scores. When there are unwanted edges present within the semantic boundary, it tends to suppress the class activation scores on either side of the gradient line, resulting in deteriorated results.

To mitigate this problem, we employ the semantic boundary in our proposed branch as a guidance for the smoothness loss. Thus, it is able to diffuse the high scores of the most discriminative area to all pixels enclosed by the semantic boundary to achieve 
smoothness. We define our semantic boundary guided smoothness loss with both first-order ($\delta_{d}$) derivative and second-order ($\delta_{d}^{2}$) derivative as:
\begin{equation}
    \mathcal{L}_{S}^{1} = \sum_{i=1}^{k} Y_{i} \sum_{u,v}^{W,H} \sum_{d \in \Vec{x}, \Vec{y}} \Psi(|\delta_{d}C_{u,v}^{i}| \exp^{-\alpha |\delta_{d}S_{u,v}^{i}|}),
\end{equation}
\begin{equation}
    \mathcal{L}_{S}^{2} = \sum_{i=1}^{k} Y_{i} \sum_{u,v}^{W,H} \sum_{d \in \Vec{x}, \Vec{y}} \Psi(|\delta_{d}^{2}C_{u,v}^{i}| \exp^{-\alpha |\delta_{d}^{2}S_{u,v}^{i}|}),
\end{equation}
where $Y \in \{0, 1\}$ is the image label to filter irrelevant classes, k is the total number of classes plus one for background, $C_{u,v}^{i}$ is the classification score at pixel $(u, v)$ for class $i$, $S_{u,v}^{i} \in \{0, 1\}$ is the semantic boundary response at pixel $(u, v)$ for class $i$, $\alpha$ is a constant set to 10. 
Function $\Psi$ is defined as $\Psi(s) = \sqrt{s^{2} + 1\exp^{-6}}$ to avoid calculating the square root of zero, following the settings in \cite{wang2018occlusion, godard2017unsupervised}, $d$ indicates that $\delta_{d}$ computes the partial derivatives in the $\Vec{x}$ or $\Vec{y}$ direction.
This semantic boundary guided smoothness loss enforces score consistency by penalising the gradients in class activation maps, with a boundary-aware term ($\delta S$ or $\delta^2 S$) to maintain the score contrast along the semantic boundary.

Then our semantic boundary guided smoothness loss function $\mathcal{L}_{S}$ is defined as weighted sum of $\mathcal{L}_{S}^{1}$ and $\mathcal{L}_{S}^{2}$ as:
\begin{equation}
\label{smoothness_loss}
    \mathcal{L}_{S} = \mathcal{L}_{S}^{1} + \lambda_{S}\mathcal{L}_{S}^{2}
\end{equation}
where $\lambda_{s}$ is a weight to balance the contribution of first-order and second-order derivative based smoothness loss, and empirically we set $\lambda_{s}= 10$, following 
settings in \cite{wang2018occlusion, godard2017unsupervised}. 

\noindent\textbf{Objective Function:} The overall loss function is composed of the baseline model loss functions, $\mathcal{L}_b$, the cross-entropy loss $\mathcal{L}_{B}$ for the training of SBDM, and the semantic boundary guided smoothness loss $\mathcal{L}_{S}$. It is defined as:
\begin{equation}
\begin{aligned}
    \mathcal{L} = &\mathcal{L}_{b}(C, Y) + \lambda_{1} \mathcal{L}_{B}(B, (S, Y)) \\
    &+ \lambda_{2} \mathcal{L}_{S}(B, (S, Y))
\end{aligned}
\end{equation}
\noindent
where $\lambda_{i}(i=1,2)$ are factors given to the semantic boundary loss and the smoothness losses. The factor of loss functions associated with the base model is kept as 1, We empirically set $\lambda_{1} = 0.05$, $\lambda_{2} = 1$. $\lambda_{1}$ is kept so that the semantic boundary loss $\mathcal{L}_{B}$ does not dominate. The choice of $\lambda_{2}$ is discussed in detail in Sec \ref{Ablation study section}.

\section{Experimental Results}
\subsection{Setup} 
\noindent\textbf{Dataset} We follow the standard procedure of WSSS to evaluate our proposed smoothness branch on the PASCAL VOC 2012 segmentation dataset \cite{everingham2010pascal} which contains 20 foreground classes and 1  background class. The official split contains 1,464 images in the training set, 1449 images in the validation set and 1456 images in the testing set. Following the common protocol for semantic segmentation, we use the augmented training set, which includes 10,582 images, provided by \cite{hariharan2011semantic}. The mean Intersection-over-Union (mIoU) metric is adopted to evaluate the segmentation results. We follow the previous WSSS works \cite{wang2020self, fan2020learning, ahn2018learning} to report results on training and validation sets for stage-1 and stage-2 models and report the results on both validation and test sets for the final semantic segmentation model.

\noindent\textbf{Training Details}
We adopt SEAM \cite{wang2020self} as our baseline model which has a ResNet38 backbone. The semantic boundary is derived from final semantic segmentation results so that we do not use any additional ground truth either implicitly or explicitly. We follow the settings of SEAM by randomly rescaling the images in the range of $[448, 768]$ along the longest side and then crop the size to $448 \times 448$. The learning rate adopts the poly policy $l_{itr} = l_{init}(1 - \frac{itr}{max.itr})^{\gamma}$, where the initial learning rate is set to $l_{init} = 0.01$ and decay is set to $\gamma = 0.9$. We initialise our model with the released weights from our baseline \cite{wang2020self}. The network is trained on 4 NVIDIA RTX 2080TI GPUs.  As our work focuses primarily on improving the quality of initial class activation maps, we just use models with publicly available Pytorch codes for the remaining stages of the pipeline.

\subsection{Comparison with the Baseline}
\noindent\textbf{Quantitative Analysis of CAM:} We compare our proposed model to the baseline model and show the corresponding performance in Table \ref{Table: compare to baseline}. Note that, unless otherwise stated, the results are evaluated on the PASCAL VOC 2012 training and validation sets. Table \ref{Table: compare to baseline} shows that our model (\enquote{Ours}\footnote{Ours indicates \enquote{CAM+SEAM+1EP} with our proposed SB.}) outperforms the baseline (\enquote{CAM+SEAM}) on both training and validation sets. The improvement on the validation set is higher that on the training set, which demonstrates that our method generalises 
well to unseen samples. We further prove that our performance improvement is not a result of additional training (an extra epoch of training). We train the baseline model with
one extra epoch of training and show its performance as \enquote{CAM+SEAM+1EP}. Although one extra epoch of training benefits the baseline model, the gap between \enquote{CAM+SEAM+1EP} and \enquote{Ours} further illustrates effectiveness of the proposed solution. The improvements are achieved without incurring additional computational cost during the inference where the smoothing branch is not involved.

\begin{table}[htb]\centering
\caption{Comparison of our method with the baseline.}
\ra{0.9}
\resizebox{\columnwidth}{!}{%
\begin{tabular}{p{5cm}p{2cm}p{2cm}}\toprule
\hfil \multirow{2}{*}{Method} & \multicolumn{2}{c}{mIoU(\%)} \\ \cmidrule{2-3}
& \hfil Train & \hfil Val \\ \midrule
CAM & \hfil 47.43 & \hfil 47.36\\
CAM + SEAM & \hfil 55.41 & \hfil 52.54 \\
CAM + SEAM + 1 EP & \hfil 55.69 & \hfil 53.06 \\
Ours & \hfil 57.32 & \hfil 55.24\\
\bottomrule
\end{tabular}
}
\label{Table: compare to baseline}
\end{table}

\begin{figure}[htb]
\small
    \centering
    \includegraphics[width=\columnwidth]{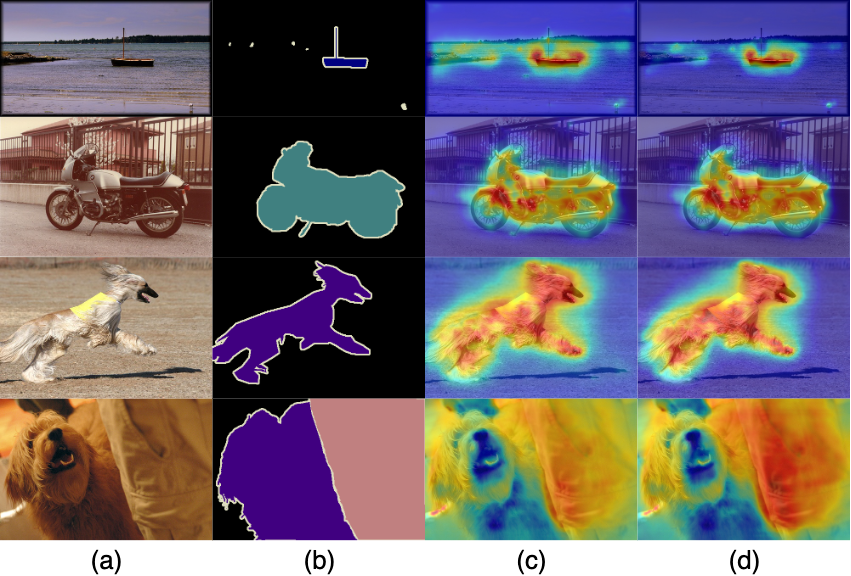}
    \vspace*{-0.5cm}
    \caption{Visualisation of class activation maps of the baseline model and our model on the PASCAL VOC 2012 validation set. (a) Original image; (b) Segmentation ground truth; (c) Baseline \cite{wang2020self}; (d) Ours. (Red represents high scores while blue represents low scores.)}
    \label{fig: comparison to baseline}
\end{figure}

\noindent\textbf{Qualitative Analysis of CAM:} To verify the effectiveness of our proposed module and loss function, we compare the class activation maps of our method with that of the baseline in Fig \ref{fig: comparison to baseline}. As the baseline model does not include structural constraints, although its class activation maps are capable of localising the objects, they are prone to over-segmentation problem. As can be seen in the boat image, it wrongly associates some water and shore area to the boat class. Our method is able to alleviate this problem by enforcing a class score gradient at the semantic boundary so that class scores at pixels outside the object area are kept low. Another advantage of our method is that the class scores within the object area are more consistent. Our method produces larger high-score areas covering the majority of the object while that of our baseline model are more sparse.

\noindent\textbf{Quantitative Analysis of the Performance after Random Walk:} We employ 
AffinityNet \cite{ahn2018learning} to learn an affinity matrix in order to perform a random walk on our produced class activation maps. As AffinityNet only utilises the pixels with high class activation scores while disregarding the low-score pixels, it is important to have class activation maps obtain consistently high class activation scores within the object area and reduce the over-segmentation on irrelevant pixels. Tab \ref{Table: compare to baseline after affinitynet} demonstrates the effectiveness of our method on the downstream random walk processing, which indicates a further 3\% performance improvement on the validation set.

\begin{table}[htb!]\centering
\caption{Comparison of the segmentation maps obtained by using AffinityNet to perform a random walk on our predicted class activation maps.}
\ra{0.9}
\resizebox{0.8\columnwidth}{!}{%
\begin{tabular}{p{4cm}p{2cm}@{}p{2cm}@{}}\toprule
\hfil \multirow{2}{*}{Method} & \multicolumn{2}{c}{mIoU(\%)} \\ \cmidrule{2-3}
& \hfil Train & \hfil Val \\ \midrule
\hfil Baseline & \hfil 63.61 & \hfil 60.08 \\
\hfil Ours & \hfil 65.31 & \hfil 63.01\\
\bottomrule
\end{tabular}
}
\label{Table: compare to baseline after affinitynet}
\end{table}

\begin{table*}[htb]\centering
\caption{Class-wise performance comparison between the pseudo ground truth of our method and our baseline on the PASCAL VOC 2012 validation set using the mIoU evaluation metric.}
\Large
\ra{0.9}
\resizebox{2\columnwidth}{!}{%
\begin{tabular}{p{3cm}p{1cm}p{1cm}p{1cm}p{1cm}p{1cm}p{1.5cm}p{1cm}p{1cm}p{1cm}p{1.2cm}p{1cm}p{1cm}p{1cm}p{1.2cm}p{1cm}p{1.5cm}p{1.2cm}p{1.2cm}p{1cm}p{1cm}p{1cm}@{}p{1.5cm}@{}}\toprule
Model & \hfil bkg & \hfil aero & \hfil bike & \hfil bird & \hfil boat & \hfil bottle & \hfil bus & \hfil car & \hfil cat & \hfil chair & \hfil cow & \hfil table & \hfil dog & \hfil horse & \hfil mbk & \hfil person & \hfil plant & \hfil sheep & \hfil sofa & \hfil train & \hfil tv & \hfil mIoU\\ \midrule

Baseline \cite{wang2020self} & \hfil 85.3 & \hfil 60.2 & \hfil 30.5 & \hfil 70.2 & \hfil 34.6 & \hfil 56.3 & \hfil 70.1 & \hfil 68.0 & \hfil 76.4 & \hfil 27.6 & \hfil \textbf{71.9} & \hfil \textbf{50.5} & \hfil \textbf{78.2} & \hfil 73.8 & \hfil 69.3 & \hfil 71.9 & \hfil 32.8 & \hfil 71.2 & \hfil 55.8 & \hfil 59.8 & \hfil 47.2 & \hfil 60.1\\

\textbf{Ours} & \hfil \textbf{87.9} & \hfil \textbf{68.5} & \hfil \textbf{32.2} & \hfil \textbf{73.6} & \hfil \textbf{48.3} & \hfil \textbf{59.4} & \hfil \textbf{75.0} & \hfil \textbf{75.2} & \hfil \textbf{77.0} & \hfil \textbf{29.4} & \hfil 64.5 & \hfil 44.6 & \hfil 76.8 & \hfil \textbf{74.3} & \hfil \textbf{70.0} & \hfil \textbf{72.9} & \hfil \textbf{45.1} & \hfil \textbf{79.7} & \hfil \textbf{56.5} & \hfil \textbf{61.0} & \hfil \textbf{51.7} & \hfil \textbf{63.0}\\
\bottomrule
\end{tabular}}
\label{Table: 20class after AffinityNet}
\end{table*}

\begin{figure}[htb]
\small
    \centering
    \includegraphics[width=\columnwidth]{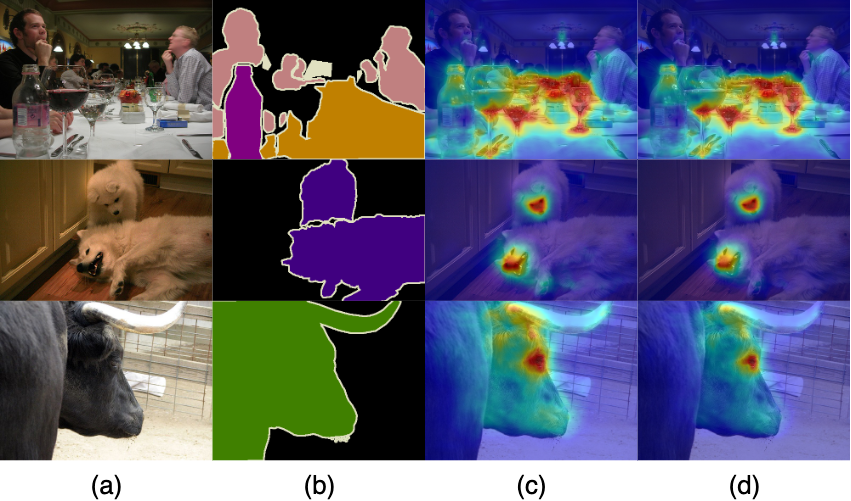}
    \vspace*{-0.5cm}
    \caption{Visualisation of failed classes on the PASCAL VOC 2012 validation set. (a) Original image; (b) Segmentation ground truth; (c) Baseline \cite{wang2020self}; (d) Ours. (Red represents high scores while blue represents low scores.)}
    \label{fig: failures}
\end{figure}

We further compare class-wise performance after the random walk in Tab \ref{Table: 20class after AffinityNet}. It shows that our method achieves consistently higher performance except on \textit{cow}, \textit{diningtable} and \textit{dog} classes. There are two main causes: (1) The initial localisation of the \textit{diningtable} class focuses primarily on the tableware rather than on the diningtable itself. Our smoothness loss cannot spread the high scores to cover the table object as shown in the first row of Fig \ref{fig: failures}; (2) \textit{Cow} and \textit{dog} classes sometimes have a single source of high-score area in the initial localisation. Its attention is placed solely on the eye and nose for the \textit{cow} and \textit{dog} class respectively. Our smoothness loss tends to find a trivial solution by creating a score contrast around these small high-score areas, leading to under-segmentation in this case. This is illustrated in the second and the third rows of Fig \ref{fig: failures}.

\begin{figure}[htb]
    \centering
    \includegraphics[width=\columnwidth]{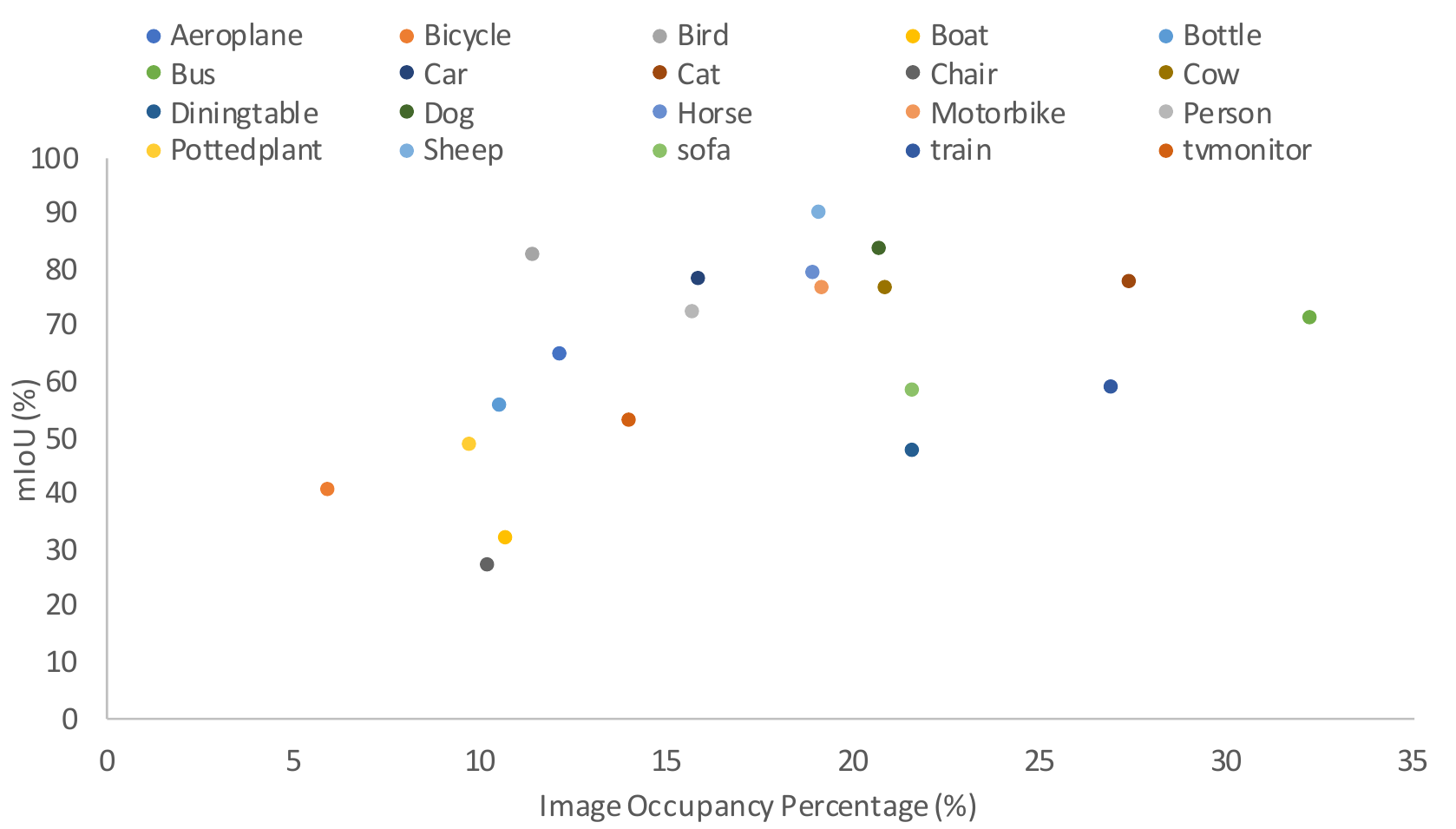}
    \vspace*{-0.5cm}
    \caption{Relationship between the mIoU and IOP for each class in the PASCAL VOC 2012 dataset.)}
    \label{fig: size}
\end{figure}

\begin{table}[h!]\centering
\caption{Ablation: components analysis of our framework.}
\large
\ra{0.9}
\resizebox{0.8\columnwidth}{!}{%
\begin{tabular}{p{2cm}p{1.5cm}p{2.5cm}p{2cm}@{}p{2cm}@{}}\toprule
\hfil \multirow{2}{*}{baseline} & \hfil \multirow{2}{*}{SBDM} & \hfil \multirow{2}{*}{Smoothness} & \multicolumn{2}{c}{mIoU(\%)} \\ \cmidrule{4-5}
& & & \hfil Train & \hfil Val \\ \midrule
\hfil \checkmark & & & \hfil 55.41 & \hfil 52.54 \\
\hfil \checkmark & \hfil \checkmark & & \hfil 55.87 & \hfil 53.35 \\
\hfil \checkmark & \hfil \checkmark & \hfil \checkmark & \hfil 57.32 & \hfil 55.24 \\
\bottomrule
\end{tabular}
}
\label{Table: ablation study}
\end{table}

\noindent\textbf{Accuracy vs Image Occupancy Percentage (IOP)} We have noticed that the mismatch between the class-wise IOP for ImageNet and PASCAL VOC 2012 results in reduced segmentation accuracy for classes that are too large or too small. IOPs of the 20 classes of PASCAL VOC 2012 dataset are between 5\% and 35\% as shown in Fig.~\ref{fig: size} while those in ImageNet are around 25\%. As a result, the group of classes with around 20\% IOP have the highest mIoU accuracy. Among this group, \textit{Diningtable} and \textit{Sofa} classes have lower mIoU because they are generally associated with heavy occlusion and potentially inaccurate initial localisation as manifested in Fig \ref{fig: failures}. Besides, the \textit{furniture} related images account for only 1.4\% of the total images in the ImageNet dataset, which is a much smaller proportion than those of other classes. \textit{Cat}, \textit{Car} and \textit{Bus} classes have around 30\% IOP. They usually have the under-segmentation problem, resulting in lower mIoU accuracy. The classes with IOP smaller than 11\% have the lowest segmentation accuracy. The mismatch on the IOP between two datasets leads to over-segmentation, which is detrimental to small class objects using mIoU evaluation metrics. The \textit{bird} class achieves accurate segmentation results despite its small IOP. This can be attributed to the fact that ImageNet has abundant images of different kinds of birds. In total, the birds images account for around 5.7\% of total images in ImageNet.\\

\subsection{Ablation Study} \label{Ablation study section}
The improvements in performance can be attributed to the structural constraints imposed by our proposed module and loss function. Table \ref{Table: ablation study} shows an ablation study of our SBDM and the semantic boundary guided smoothness loss function. Through the SBDM, the feature maps from multiple levels of the encoder become structure-aware, which indirectly enhances the quality of the final score map by passing structural information. The major improvements come from the semantic boundary-guided smoothness loss function. It creates a sharp contrast of class activation scores around the semantic boundary while enforcing more consistent high scores within the area enclosed by the semantic boundary.

\noindent\textbf{Semantic Boundary Detection Module:} Our SBDM uses both feature maps $F_{i}(i = 1, 3, 5, 6)$ and Canny edges to predict the semantic boundary. SDRG \cite{huang2018weakly} has also used boundary information to impose structural constraints, however, they only use it to separate the foreground and background pixels. We demonstrate that making feature maps structure-aware is beneficial to predicting class activation maps. Tab.~\ref{Table: SBDM} shows that SBDM requires both high-level features and low-level features to determine the boundary semantically and spatially. The Canny edge is demonstrated to provide extra spatial cues for the bottom-up estimation of the semantic boundary, leading to improved performance.

\begin{table}[h!]\centering
\caption{Ablation: components analysis of the semantic boundary detection module.}
\large
\ra{0.9}
\resizebox{0.8\columnwidth}{!}{%
\begin{tabular}{p{1.5cm}p{1cm}p{1cm}p{1cm}p{1cm}p{2cm}@{}p{2cm}@{}}\toprule
\hfil \multirow{2}{*}{Canny} & \hfil \multirow{2}{*}{$F_{1}$} & \hfil \multirow{2}{*}{$F_{3}$} & \hfil \multirow{2}{*}{$F_{5}$} & \hfil \multirow{2}{*}{$F_{6}$} & \multicolumn{2}{c}{mIoU(\%)} \\ \cmidrule{6-7}
& & & & & \hfil Train & \hfil Val \\ \midrule
& & & & & \hfil 55.41 & \hfil 52.54 \\
 & & & \hfil \checkmark & \hfil \checkmark &  \hfil 55.17 & \hfil 52.57 \\
 & \hfil \checkmark & \hfil \checkmark & & & \hfil 55.05 & \hfil 52.46 \\
 & \hfil \checkmark & \hfil \checkmark & \hfil \checkmark & \hfil \checkmark & \hfil 55.78 & \hfil 53.31 \\
\hfil \checkmark & \hfil \checkmark & \hfil \checkmark & \hfil \checkmark & \hfil \checkmark & \hfil 55.87 & \hfil 53.35 \\
\bottomrule
\end{tabular}
}
\label{Table: SBDM}
\end{table}

\noindent\textbf{Semantic Boundary-Guided Smoothness Loss:} 
The semantic boundary is essential for the smoothness loss to be used in WSSS with only image labels. As unwanted noisy edges are not present within the semantic boundary, it can enlarge the high-score area while alleviating 
over-segmentation. Tab.~\ref{Table: smoothness factor} demonstrates that the factor $\lambda_{2}$ assigned to the smoothness loss has a big impact on its performance. When the factor is too low, the smoothness loss fails to spread to high scores to surrounding pixels. On the contrary, when 
the factor is 
too high, it searches for a trivial solution by reducing all scores to minimal values, resulting in highly deteriorated results.

\begin{table}[h!]\centering
\caption{Analysis of the contribution of semantic boundary-guided smoothness loss to our performance.}
\ra{0.9}
\resizebox{0.8\columnwidth}{!}{%
\begin{tabular}{p{3cm}p{2cm}@{}p{2cm}@{}}\toprule
\hfil \multirow{2}{*}{$\lambda_{2}$} & \multicolumn{2}{c}{mIoU(\%)} \\ \cmidrule{2-3}
& \hfil Train & \hfil Val \\ \midrule
\hfil 0.5 & \hfil 56.75 & \hfil 54.48 \\
\hfil 0.75 & \hfil 57.25 & \hfil 54.99\\
\hfil 1.0 & \hfil 57.32 & \hfil 55.24 \\
\hfil 1.25 & \hfil 57.25 & \hfil 55.22\\
\hfil 1.5 & \hfil 56.40 & \hfil 54.52 \\
\hfil 2.0 & \hfil 34.98 & \hfil 33.36\\
\bottomrule
\end{tabular}
}
\label{Table: smoothness factor}
\end{table}

\subsection{Comparison with WSSS state-of-the-art}
\begin{figure*}[htb!]
    \centering
    \includegraphics[width=1.95\columnwidth]{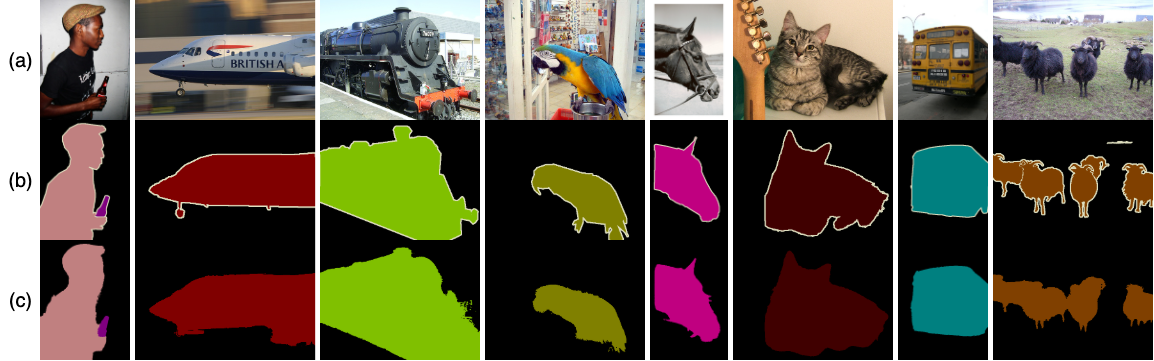}
    \vspace*{-0cm}
    \caption{Visualisation of semantic segmentation results on the PASCAL VOC 2012 validation set. (a) Original image; (b) Segmentation ground truth; (c) Ours.}
    \label{fig: deeplan visualisation}
\end{figure*}

Tab.~\ref{Table: deeplab with image-level labels} illustrates the results of previous WSSS methods with only image-level labels and ours and Tab.~\ref{Table: 20class} provides more detailed comparison on class-wise performance. Our method improves upon our baseline model on both validation and test sets. Furthermore, when no additional supervision is
used either implicitly or explicitly, our method achieves the state-of-the-art results on the PASCAL VOC 2012 test set. As WSSS with image-level labels does not explicitly predict the background class, many methods separate 
foreground pixels from 
background ones by leveraging an external saliency detector, which implicitly uses saliency ground truth as additional supervision 
which includes object segmentation boundary information. Tab.~\ref{Table: deeplab with image-level labels and external saliency detector} shows previous WSSS methods equipped with an external saliency detector. It can be seen that our method can still outperform most of them without implicitly using the saliency ground truth except the ICD method \cite{fan2020learning}. We also visualise some qualitative results of our semantic segmentation model in Fig.~\ref{fig: deeplan visualisation}, which demonstrates its ability to segment objects belonging to different classes.

\begin{table}[htb!]\centering
\caption{Performance comparison of our method with previous WSSS state-of-the-art with only image-level labels on PASCAL VOC 2012 using the mIoU evaluation metric.}
\ra{0.9}
\resizebox{0.9\columnwidth}{!}{%
\begin{tabular}{p{3cm}p{2cm}p{1.5cm}@{}p{1.5cm}@{}}\toprule
\hfil \multirow{2}{*}{Methods} & \hfil \multirow{2}{*}{Backbone} & \multicolumn{2}{c}{mIoU(\%)} \\ \cmidrule{3-4}
& & \hfil Val & \hfil Test \\ \midrule
CCNN \cite{pathak2015constrained} & \hfil VGG16 & \hfil 35.3 & \hfil 35.6 \\
EM-Adapt \cite{papandreou2015weakly} & \hfil VGG16 & \hfil 38.2 & \hfil 39.6 \\
MIL \cite{pinheiro2015image} & \hfil VGG16 & \hfil 42.0 & \hfil 43.2 \\
SEC \cite{kolesnikov2016seed} & \hfil VGG16 & \hfil 50.7 & \hfil 51.1 \\
AugFeed \cite{qi2016augmented} & \hfil VGG16 & \hfil 54.3 & \hfil 55.5 \\
AdvErasing \cite{wei2017object} & \hfil VGG16 & \hfil 55.0 & \hfil 55.7 \\
AffinityNet \cite{ahn2018learning} & \hfil ResNet38 & \hfil 61.7 & \hfil 63.7 \\
SSNet \cite{araslanov2020single} & \hfil ResNet38 & \hfil 62.7 & \hfil 64.3 \\
ICD \cite{fan2020learning} & \hfil ResNet101 & \hfil 64.1 & \hfil 64.3\\
IRNet \cite{ahn2019weakly} & \hfil ResNet50 & \hfil 63.5 & \hfil 64.8 \\
SSENet \cite{wang2019self} & \hfil ResNet38 & \hfil 63.3 & \hfil 64.9 \\
SEAM \cite{wang2020self} & \hfil ResNet38 & \hfil 64.5 & \hfil 65.7\\
SC-CAM \cite{chang2020weakly} & \hfil ResNet101 & \hfil 66.1 & \hfil 65.9\\\midrule
\textbf{Ours} & \hfil ResNet101 & \hfil \textbf{66.5} & \hfil \textbf{66.7}\\
\bottomrule
\end{tabular}
}
\label{Table: deeplab with image-level labels}
\end{table}

\begin{table}[htb!]\centering
\caption{Results of previous WSSS state-of-the-art methods with image-level labels and external saliency detector on PASCAL VOC 2012 using the mIoU evaluation metric.}
\ra{0.9}
\resizebox{0.9\columnwidth}{!}{%
\begin{tabular}{p{3cm}p{2cm}p{1.5cm}@{}p{1.5cm}@{}}\toprule
\hfil \multirow{2}{*}{Methods} & \hfil \multirow{2}{*}{Backbone} & \multicolumn{2}{c}{mIoU(\%)} \\ \cmidrule{3-4}
& & \hfil Val & \hfil Test \\ \midrule
STC \cite{wei2016stc} & \hfil VGG16 & \hfil 49.8 & \hfil 51.2 \\
MDC \cite{wei2018revisiting} & \hfil VGG16 & \hfil 60.4 & \hfil 60.8 \\
MCOF \cite{wang2018weakly} & \hfil ResNet101 & \hfil 60.3 & \hfil 61.2\\
DCSP \cite{chaudhry2017discovering} & \hfil ResNet101 & \hfil 60.8 & \hfil 61.9\\
SeeNet \cite{hou2018self} & \hfil ResNet101 & \hfil 63.1 & \hfil 62.8 \\
DSRG \cite{huang2018weakly} & \hfil ResNet101 & \hfil 61.4 & \hfil 63.7 \\
FickleNet \cite{lee2019ficklenet} & \hfil ResNet101 & \hfil 64.9 & \hfil 65.3 \\
OAA \cite{jiang2019integral} & \hfil ResNet101 & \hfil 65.2 & \hfil 66.4\\
ICD \cite{fan2020learning} & \hfil ResNet101 & \hfil 67.8 & \hfil 68.0\\
\bottomrule
\end{tabular}
}
\label{Table: deeplab with image-level labels and external saliency detector}
\end{table}

\section{Conclusion} In this paper, we propose a Smoothing Branch (SB) to impose structural constraints in the training of WSSS models with image-level labels. The SB leverages the derived semantic boundary to make the feature maps structure-aware, which subsequently improves the generated class activation maps. The semantic boundary also guides the smoothness loss function to make the scores more consistent within the enclosed object area and reduce the scores at outside pixels. Comprehensive experiments are conducted to study the components of our proposed method. The generated class activation maps are thus capable of better preserving the object structure. The semantic segmentation network trained with our segmentation maps as pseudo ground truth achieves state-of-the-art performance on the PASCAL VOC 2012 dataset, demonstrating the superior performance of our method.

\begin{table*}[htb!]\centering
\caption{Class-wise performance comparison of our method with previous WSSS state-of-the-art methods on the PASCAL VOC 2012 validation set using the mIoU evaluation metric.}
\Large
\ra{0.9}
\resizebox{2\columnwidth}{!}{%
\begin{tabular}{p{3.5cm}p{1cm}p{1cm}p{1cm}p{1cm}p{1cm}p{1.5cm}p{1cm}p{1cm}p{1cm}p{1.2cm}p{1cm}p{1cm}p{1cm}p{1.2cm}p{1cm}p{1.5cm}p{1.2cm}p{1.2cm}p{1cm}p{1cm}p{1cm}@{}p{1.5cm}@{}}\toprule
Model & \hfil bkg & \hfil aero & \hfil bike & \hfil bird & \hfil boat & \hfil bottle & \hfil bus & \hfil car & \hfil cat & \hfil chair & \hfil cow & \hfil table & \hfil dog & \hfil horse & \hfil mbk & \hfil person & \hfil plant & \hfil sheep & \hfil sofa & \hfil train & \hfil tv & \hfil mIoU\\ \midrule

CCNN \cite{pathak2015constrained} & \hfil 68.5 & \hfil 25.5 & \hfil 18.0 & \hfil 25.4 & \hfil 20.2 & \hfil 36.3 & \hfil 46.8 & \hfil 47.1 & \hfil 48.0 & \hfil 15.8 & \hfil 37.9 & \hfil 21.0 & \hfil 44.5 & \hfil 34.5 & \hfil 46.2 & \hfil 40.7 & \hfil 30.4 & \hfil 36.3 & \hfil 22.2 & \hfil 38.8 & \hfil 36.9 & \hfil 35.3\\

MIL \cite{pinheiro2015image} & \hfil 79.6 & \hfil 50.2 & \hfil 21.6 & \hfil 40.9 & \hfil 34.9 & \hfil 40.5 & \hfil 45.9 & \hfil 51.5 & \hfil 60.6 & \hfil 12.6 & \hfil 51.2 & \hfil 11.6 & \hfil 56.8 & \hfil 52.9 & \hfil 44.8 & \hfil 42.7 & \hfil 31.2 & \hfil 55.4 & \hfil 21.5 & \hfil 38.8 & \hfil 36.9 & \hfil 42.0\\

SEC \cite{kolesnikov2016seed} & \hfil 82.4 & \hfil 62.9 & \hfil 26.4 & \hfil 61.6 & \hfil 27.6 & \hfil 38.1 & \hfil 66.6 & \hfil 62.7 & \hfil 75.2 & \hfil 22.1 & \hfil 53.5 & \hfil 28.3 & \hfil 65.8 & \hfil 57.8 & \hfil 62.3 & \hfil 52.5 & \hfil 32.5 & \hfil 62.6 & \hfil 32.1 & \hfil 45.4 & \hfil 45.3 & \hfil 50.7\\

AdvErasing \cite{wei2017object} & \hfil 83.4 & \hfil 71.1 & \hfil 30.5 & \hfil 72.9 & \hfil 41.6 & \hfil 55.9 & \hfil 63.1 & \hfil 60.2 & \hfil 74.0 & \hfil 18.0 & \hfil 66.5 & \hfil 32.4 & \hfil 71.7 & \hfil 62.0 & \hfil 64.8 & \hfil 52.4 & \hfil 37.4 & \hfil 69.1 & \hfil 31.4 & \hfil 58.9 & \hfil 43.9 & \hfil 55.0\\

AffinityNet \cite{ahn2018learning} & \hfil 88.2 & \hfil 68.2 & \hfil 30.6 & \hfil 81.1 & \hfil 49.6 & \hfil 61.0 & \hfil 77.8 & \hfil 66.1 & \hfil 75.1 & \hfil 29.0 & \hfil 66.0 & \hfil 40.2 & \hfil 80.4 & \hfil 62.0 & \hfil 70.4 & \hfil 73.7 & \hfil 42.5 & \hfil 70.7 & \hfil 42.6 & \hfil 68.1 & \hfil 51.6 & \hfil 61.7\\

SEAM \cite{wang2020self} & \hfil 88.8 & \hfil 68.5 & \hfil \textbf{33.3} & \hfil \textbf{85.7} & \hfil 40.4 & \hfil \textbf{67.3} & \hfil 78.9 & \hfil 76.3 & \hfil 81.9 & \hfil 29.1 & \hfil \textbf{75.5} & \hfil \textbf{48.1} & \hfil \textbf{79.9} & \hfil 73.8 & \hfil 71.4 & \hfil 75.2 & \hfil \textbf{48.9} & \hfil 79.8 & \hfil \textbf{40.9} & \hfil 58.2 & \hfil 53.0 & \hfil 64.5\\

\midrule

Ours & \hfil \textbf{89.6} & \hfil \textbf{73.0} & \hfil 33.0 & \hfil 83.5 & \hfil \textbf{54.0} & \hfil67.6 & \hfil \textbf{84.1} & \hfil \textbf{80.7} & \hfil \textbf{84.4} & \hfil \textbf{30.3} & \hfil 75.3 & \hfil 43.7 & \hfil 78.0 & \hfil \textbf{79.7} & \hfil \textbf{74.1} & \hfil \textbf{75.8} & \hfil 46.6 & \hfil \textbf{84.3} & \hfil 37.0 & \hfil \textbf{61.4} & \hfil \textbf{58.5} & \hfil \textbf{66.5}\\

\bottomrule
\end{tabular}
}
\label{Table: 20class}
\end{table*}

\printbibliography

\end{document}